# Multi-objects association in perception of dynamical situation

**Dominique GRUYER, Véronique BERGE-CHERFAOUI**
HeuDiaSyC-UMR CNRS 6599
Université de Technologie de Compiègne
BP 20 529, 60205 Compiègne, France
e-mail :gruyer, vberge@hds.utc.fr

## Abstract

In current perception systems applied to the rebuilding of the environment for intelligent vehicles, the part reserved to object association for the tracking is increasingly significant. This allows firstly to follow the objects temporal evolution and secondly to increase the reliability of environment perception. We propose in this communication the development of a multi-objects association algorithm with ambiguity removal entering into the design of such a dynamic perception system for intelligent vehicles.
This algorithm uses the belief theory and data modelling with fuzzy mathematics in order to be able to handle inaccurate as well as uncertain information due to imperfect sensors. These theories also allow the fusion of numerical as well as symbolic data. We develop in this article the problem of matching between known and perceived objects. This makes it possible to update a dynamic environment map for a vehicle. The belief theory will enable us to quantify the belief in the association of each perceived object with each known object. Conflicts can appear in the case of object appearance or disappearance, or in the case of a confused situation or bad perception. These conflicts are removed or solved using an assignment algorithm, giving a solution called the « best » and so ensuring the tracking of some objects present in our environment.

## 1 Introduction

Our research is focused on perception systems for vehicle in road situations. Due to the increasing number of road vehicles, many security problems are appearing. In order to solve these problems, the European project Prometheus [ROMB 93] has given the beginning of an answer about what could be the vehicle of the future: an interaction between the driver and the vehicle through a driving assistance system.

Our work is carrying on this project. We are particularly interested in data fusion algorithms giving the driver some accurate, especially reliable and pertinent information in relation with the current situation. In order to deal with these security problems and information reliability, we have designed a perception algorithm combining some tools dealing simultaneously with the inaccuracy and uncertainty of dynamic environment representation.

In this communication, we present a multi-object matching algorithm with ambiguity removal. Its goal is to associate perceived objects with known objects using fuzzy measures (high level data) and fuzzy prediction windows obtained from a fuzzy estimator-predictor presented in [GRUY 98a][GRUY 98b]. This algorithm uses the belief theory and data modelling with fuzzy mathematics in order to be able to handle inaccurate as well as uncertain information due to imperfect sensors. These theories also allow the fusion of numerical as well as symbolic data. The fuzzy measures represent the perceived objects and the fuzzy prediction windows represent the known objects. This matching algorithm makes it possible to pass from a multi-target detection mode to a multi-target tracking mode. This tracking mode gives the possibility to take into account the appearance and disappearance of every object within our environment.

We quickly present, in a first part, the basic notions, the general points and the disadvantages of the belief theory. We then suggest a generalisation of the Dempster's combination rule [SHAF 76] applied to our problem. These works are based on Michèle Rombaut research [ROMB 98]. We will see that the matching may create a conflicting decision. In a second part, we will give an optimal solution to remove these conflicts and then obtain a new decision called « the best » according to a decision criterion developed later. Afterwards, we will describe a way to build the initial mass set through a concordance operator between known and perceived objects. We will finish with some operating examples and a succinct presentation of the object propagation stage which makes the temporal tracking of all objects with dynamical uncertainty management in time possible. Then we will conclude and present our future works.

## 2   Belief theory for dynamic association

### 2.1   Generalities

Belief theory allows both to model and to use uncertain and inaccurate data, as well as qualitative and quantitative data, so as to keep a consistency and homogeneity with all concepts and tools developed in the remaining of our algorithm shown in [GRUY 98b].

In a general framework, we can say that our problem consists in identifying an object designated by a generic variable $X$ among a set of hypotheses $Y_i$. One of these hypotheses is in a position to be the solution. In our case, we want to associate perceived objects $X_i$ to known objects $Y_j$. Belief theory enable us to value the veracity of $P_i$ propositions representing the matching of our different objects. These propositions can be simple as well as complex.

**Example:**

$P_1$ = " perceived object $X$ is known object $Y_i$ "

$P_2$ = " perceived object $X$ is known objects $Y_i$ or $Y_j$ "

We must then define a magnitude allowing the characterization of this truth. This magnitude is the elementary probabilistic mass $m_\Theta()$ defined on [0,1]. This mass is very near to the probabilistic mass, to the exception that we do not share this mass only on single elements but on all elements of the definition referential $2^\Theta = \{A/A \subseteq \Theta\} = \{\varnothing, Y_1, Y_2,..., Y_n, Y_1 \cup Y_2,...,\Theta\}$. This referential is build through the frame of discernment $\Theta = \{Y_1, Y_2, \cdots, Y_n\}$, which regroups all admissible hypotheses, these hypotheses having to be exclusive. ($Y_i \cap Y_j = \varnothing$, $\forall$ i $\neq$ j). This distribution is a function of the knowledge about the source to model. The whole mass obtained is called « distribution of mass ». The sum of these masses is equal to 1 and the mass given to impossible case m($\varnothing$) must be equal to 0.

### 2.2   Generalised combination and multi-object association

The combination of information coming from different sources has the advantage to increase the information reliability and to reduce the influence of failing information (inaccurate, uncertain, incomplete and conflicting). But to obtain this result, it's necessary to have complementary and/or redundant information.

We combine these pieces of information with the Dempster-Shafer combination rule. This rule applied on $n$ sources gives a combination in series:

$$m_\Theta = m_\Theta^{S_1} \oplus \cdots \oplus m_\Theta^{S_n} = ((((m_\Theta^{S_1} \oplus m_\Theta^{S_2}) \oplus m_\Theta^{S_3}) \oplus \cdots) \oplus m_\Theta^{S_n})$$

In the framework of a processing in « *closed-world* », that is with an exhaustive frame of discernment the combination of a great number of sources lead to a combinatorial explosion. This is the main drawback of this combination rule. On the other hand, it offers the advantage of being associative and commutative, which is not the case of the majority of the fusion operators [BLOC 96].

In order to succeed in generalizing the Dempster combination rule and thus reducing its combinative complexity, we will limit the reference frame of definition while adding as a constraint that a perceived object can be connected with one and only one known object.

For example, for a detected object to associate among three known objects, we will have the following frame of discernment:

$$\Theta = \{Y_1, Y_2, Y_3\}$$

with $Y_i$ meaning $X$ is in relation with $Y_i$

From this frame of discernment, we build the referential of definition according to:

$$2^\Theta = \{*, Y_1, Y_2, Y_3, Y_1 \cup Y_2, Y_1 \cup Y_3, Y_2 \cup Y_3, \Theta\}$$

$$2^\Theta = \{*, Y_1, Y_2, Y_3, \overline{Y_1}, \overline{Y_2}, \overline{Y_3}, \Theta\}$$

with $\overline{Y_i}$, meaning $X$ is not in relation with $Y_i$

In this referential of definition, we find the singleton hypotheses of the frame of discernment to which we add ignorance with the hypothesis $\Theta$ and the notion of « *empty* » or more explicitly the « *nothing* » with the hypothesis *. We obtain then a distribution of masses made up of the following masses:

$m_{i,j}(Y_j)$ :  mass associated with the proposition « $X_i$ *is in relation with $Y_j$.* »

$m_{i,j}(\overline{Y_j})$ :  mass associated with the proposition « $X_i$ *is not in relation with $Y_j$.* »

$m_{i,j}(\Theta_{i,j})$ : mass representing ignorance.

$m_{i,\cdot}(*)$ :  mass representing the reject : «$X_i$ *is in relation with nothing.*»

In this mass distribution, the first index $i$ indicates the processed perceived object and the second index $j$ the known object. If one index is replaced by a dot, it means that the mass is applied to all objects perceived or none according to the location of this dot.

Moreover, if we use a combination in cascade, the mass $m_{i,\cdot}(*)$ is not part of the initial mass set and appears only after the first combination. It replaces the conjunction of the combined masses $m_{i,j}(\overline{Y_j})$.

By observing the behaviour of the combination in cascade with $n$ mass sets, we revealed a general behavior which enables us to put in equation the final mass set according to the initial mass sets. This enables us to obtain an independence of our final masses in relation to the recurrence of the combination.

$$m_{i,\cdot}(Y_j) = K_{i,\cdot} \cdot m_{i,j}(Y_j) \cdot \prod_{\substack{k=1\cdots n \\ k \neq j}}(1-m_{i,k}(Y_k))$$

$$m_{i,\cdot}(*) = K_{i,\cdot} \cdot \prod_{j=1\cdots n} m_{i,j}(\overline{Y}_j)$$

$$m_{i,\cdot}(\Theta) = K_{i,\cdot} \cdot \left(\prod_{j=1\cdots n}[m_{i,j}(\Theta_{i,j})+m_{i,j}(\overline{Y}_j)] - \prod_{j=1\cdots n} m_{i,j}(\overline{Y}_j)\right)$$

avec $K_{i,\cdot} = \prod_{l=1\cdots n} K_{i,l}$

This $K_{i,\cdot}$ is the re-normalization of the $n$ combinations, the product of the various re-normalization carried out during all the combinations.

$$m_{i,\cdot}(\varnothing) = \left[\prod_{l=1\cdots n-1} K_{i,l}\right] \cdot m_{i,n}(Y_n) \cdot \sum_{k=1\cdots n-1}\left(m_{i,k}(Y_k) \cdot \left[\prod_{\substack{m=1\cdots n-1 \\ m \neq k}} A_m\right]\right)$$

with $A_m = m_{i,m}(\Theta_{i,m}) + m_{i,m}(\overline{Y}_m)$

$$K_{i,\cdot} = \frac{1}{1-m_{i,\cdot}(\varnothing)}$$

$$K_{i,\cdot} = \frac{1}{\prod_{j=1\cdots n}(1-m_{i,j}(Y_j)) \cdot \left(1+\sum_{j=1}^{n}\frac{m_{i,j}(Y_j)}{1-m_{i,j}(Y_j)}\right)}$$

From each mass set, we build two matrices $M_{i,\cdot}^\sigma$ and $M_{\cdot,j}^\sigma$ which give the belief that a perceived object is associated with a known object and conversely (see tab 1 and 2). With these two matrices, we can manage object appearance and disappearance. Moreover, if the two decisions obtained from these two matrices agree, then we can validate this decision without using the algorithm of ambiguity removal.

The sum of the elements of each column is equal to 1 because of the re-normalization. The resulting frames of discernment are:

$$\Theta_{\cdot,j} = \{Y_{1,j}, Y_{2,j}, \cdots, Y_{n,j}, Y_{*,j}\} \text{ and}$$
$$\Theta_{i,\cdot} = \{X_{i,1}, X_{i,2}, \cdots, X_{i,m}, X_{i,*}\}$$

We can interpret $Y_{*,j}$ by the relation «*no perceived object $X_i$ is in relation with the known object $Y_j$*» and $Y_{i,*}$ by the relation «*the perceived object $X_i$ is not dependent with any known object*». In the first case we can deduce from it that an object has just disappeared and in the second case, that an object has just appeared. These objects, which appeared or disappeared, can also be false alarms.

The following stage consists in establishing the best decision on association using the two matrices obtained previously.

**Tab 1** : Belief matrix $M_{i,\cdot}^\sigma$ quantifying the relation between perceived objects and known objects.

| $M_{i,\cdot}^\sigma$ | $X_1$ | $X_2$ | ... | $X_n$ | |
|---|---|---|---|---|---|
| $Y_1$ | $m_{1,\cdot}(Y_1)$ | $m_{2,\cdot}(Y_1)$ | | $m_{n,\cdot}(Y_1)$ | |
| $Y_2$ | $m_{1,\cdot}(Y_2)$ | $m_{2,\cdot}(Y_2)$ | | $m_{n,\cdot}(Y_2)$ | |
| ... | | | | | A1 |
| $Y_m$ | $m_{1,\cdot}(Y_m)$ | $m_{2,\cdot}(Y_m)$ | | $m_{n,\cdot}(Y_m)$ | |
| * | $m_{1,\cdot}(*)$ | $m_{2,\cdot}(*)$ | | $m_{n,\cdot}(*)$ | |
| $\Theta$ | $m_{1,\cdot}(\Theta)$ | $m_{2,\cdot}(\Theta)$ | | $m_{n,\cdot}(\Theta)$ | B1 |

**Tab 2** : Belief matrix $M_{\cdot,j}^\sigma$ quantifying the relation between known objects and perceived objects.

| $M_{\cdot,j}^\sigma$ | $Y_1$ | $Y_2$ | ... | $Y_m$ | |
|---|---|---|---|---|---|
| $X_1$ | $m_{\cdot,1}(X_1)$ | $m_{\cdot,2}(X_1)$ | | $m_{\cdot,m}(X_1)$ | |
| $X_2$ | $m_{\cdot,1}(X_2)$ | $m_{\cdot,2}(X_2)$ | | $m_{\cdot,m}(X_2)$ | |
| ... | | | | | A2 |
| $X_n$ | $m_{\cdot,1}(X_n)$ | $m_{\cdot,2}(X_n)$ | | $m_{\cdot,m}(X_n)$ | |
| * | $m_{\cdot,1}(*)$ | $m_{\cdot,2}(*)$ | | $m_{\cdot,m}(*)$ | |
| $\Theta$ | $m_{\cdot,1}(\Theta)$ | $m_{\cdot,2}(\Theta)$ | | $m_{\cdot,m}(\Theta)$ | B2 |

As we use a referential of definition built with singleton hypotheses, except $\Theta$ and *, the use of the mass redistribution function would not add any useful information. This redistribution would simply reinforce the fact that our perceived object is really in relation with a known object. This is why we use as our decision criterion the maximum of belief on each column of the two belief matrices.

$$d(Y_{i,\cdot}) = \underset{j}{Max}[M_{i,j}^{Cr}]$$

This rule answers the question « *what is the known object $Y_j$ in relation with the perceived object $X_i$* ». We have the same rule for the known objects:

$$d(X_{\cdot,j}) = \underset{i}{Max}[M_{i,j}^{Cr}]$$

The problem is then to know how to process ambiguities. Ambiguity will intervene when an object, perceived or known, is in relation with two perceived or known objects, or if the first maximization gives a decision on the relation between objects $X_i$ and $Y_j$ and the second maximization gives a decision contradictory to the first, for example $Y_j$ in relation to $X_k$, $i \neq k$.

The following step consists in obtaining a matrix in which all the objects will be classified without ambiguities and with a maximisation of the belief on the decision.

One wants to thus ensure that the decision taken is not " *good* " but " *the best* ". By the " *best* ", we mean that if we have a known object and some defective or frustrate sensor to perceive it, then we are unlikely to know what this object corresponds to, and therefore we have a little chance to ensure that the association is good. But among all the available possibilities, we must certify that the decision is the " *best* " of all possible decisions.

It is thus necessary to find a way of combining the lines and the columns of the two matrices of beliefs in order to obtain a new general matrix representing final associations. The following chapter describes how to solve this problem while avoiding the study of all the combinations of the elements concerned by the conflict.

## 3. Conflicts Resolution

### 3.1 Affectation: a solution to resolve the conflicts

In order to use the most of possible information and to obtain an optimal decision with maximisation of the sum of beliefs, we have decomposed the two belief matrix $M_{.,j}^{\sigma}$ et $M_{i,.}^{\sigma}$. Thus we obtain a more synthetic new structure combining all the information at our disposal.

The decomposition of each one of these matrices gives the sub-matrices $A_1$, $A_2$ and $B_1$, $B_2$ (see tab 1 and 2). The two first matrices represent the relations between the various objects and the two others represent the impossibility and unknown concepts on the relations linking the objects.

Tab 3: Combination of A1 and A2 matrices

| $M_{i,j}^{\sigma}$ | | | $Y_m$ | |
|---|---|---|---|---|
| | $m_{1,.}(Y_1) \cdot m_{.,1}(X_1)$ | | $m_{1,.}(Y_m) \cdot m_{.,m}(X_1)$ | |
| | | | | |
| $X_n$ | $m_{n,.}(Y_1) \cdot m_{.,1}(X_n)$ | | $m_{n,.}(Y_m) \cdot m_{.,m}(X_n)$ | |

The two matrices $B_1$ and $B_2$ contain information on appearance or disappearance of targets, or the expression of conflict. By applying a conjunction to all the relations included in the matrices $A_1$ and $A_2$, we obtain a more synthetic new matrix that represents the relations between the $n$ perceived objects and the $m$ known objects (see tab 3). This matrix is homogeneous since we handle the same objects in the two matrices we combine.

We can interpret this new matrix as being a cost matrix connecting two sets of data. Our goal is now to find the best two to two assignment of $n$ perceived objects with the $m$ known objects. Setting this matrix in graph form brings us back to a traditional problem of assignment which is generally seen as a particular case of the transport problem without capacities. It can also be seen as a problem of perfect coupling with minimum weight (or maximum) in a bipartite graph [GOND 95].

If the known objects are independent, the total belief on our coupling is the sum of the beliefs of each couple *perceived objects / known objects*. Our problem is thus an assignment problem on the bipartite graph $G=(X,Y,A)$ with $X$ the perceived objects, $Y$ the known objects and $A$ the set of arcs of this graph.

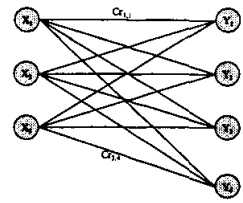

**Figure 3.1** Graph corresponding to belief matrix $M_{i,j}^{cr}$

An arc of this graph will indicate a possible assignment of a perceived object with a known object and will be valued by the corresponding belief (shown in fig 3.1). The required solution is thus a coupling of $n$ arcs and maximum belief with a constraint of not-adjacency on the couples which means that perceived object can be associated with one and only one known object and reciprocally. These algorithms of coupling have the advantage of generalising the assignment problems and being a part of a class of linear programs in integer numbers which admit an resolution algorithm with polynomial complexity in N and M (the number of arcs and the number of nodes of the graph) [LAWL 76].

In our system, we used a traditional assignment algorithm called the Hungarian algorithm [KUHN 55].

We are going to expose now the general principle of this algorithm. It is build from four main operations: **coupling, improving chains search, marking** and **admissible arcs modification.**

**Find admissible arcs:** This algorithm relies on the fact that we do not modify the problem by removing from a line (or a column) any number $\alpha$ corresponding to the smaller elements of this line (or this column). This property thus enables us to reveal the admissible arcs representing the most probable relations.

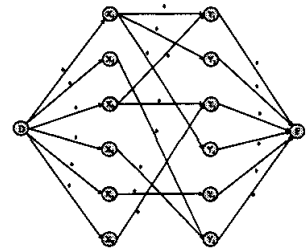

**Figure 3.2** Example of a bipartite graph built from an admissible arcs set.

This avoids an exhaustive processing of the relations between objects. This operation is obtained by revealing an ' 0 ' in each line and each column of the belief's matrix $\overline{M}_{i,j}^{\sigma}$ (with $\overline{C}_{ij} = 1 - C_{ij}$). $C_{ij}$ is the belief on one matrix belief relation.

**Coupling:** With these admissible arcs, we build a bipartite graph in which we seek the maximum cardinality coupling (see fig 3.2). This example shows us the most probable relations existing between six known objects and six perceived objects.

Once the first coupling is made, we have a subset of nonadjacent arcs marked by ' **1** ' which indicates that a relation is validated, and a second subset of arcs marked by ' 0 ' which are not used.

**Improving chains search:** The following stage will consist in finding a possible improvement of this coupling. Seeking the improving chains existing in this graph does this improvement. An improving chain is a succession of arcs alternatively valued with ' 0 ' and ' 1 ' that goes from the node $D$ to the node $F$. Once a chain is found, we carry out a transfer on this chain. This means that the marking of the arcs is reversed.

**Marking:** The existence of a set of chains that go from $D$ and are not leading to $F$ indicates that our coupling is not optimal (coupling cardinality) and thus should be improved. For that we will mark all the nodes included in these non-improving chains and thus will obtain four subsets of objects:

  $X^*$ and $Y^*$: Not marked perceived and known objects.
  $X$-$X^*$:   Marked perceived objects.
  $Y$-$Y^*$:   Marked known objects.

This marking is obtained by using the Ford-Fulkerson algorithm [PRIN 94].

We can interpret this marking as follows: Firstly, there are no admissible arcs between $X^*$ and $Y$-$Y^*$, therefore the reduction operation of the belief matrix has told us that the relations between the set of perceived objects $X^*$ and known objects $Y$-$Y^*$ are strongly improbable. No arc was generated in the graph between these two object subsets. And secondly, the arcs between $X$-$X^*$ and $Y^*$ have a null flow. That mean the perceived objects and the known objects belonging to these two subsets were not associated by the coupling algorithm.

**Admissible arcs modification:** The following stage consists in making appear, disappear or in preserving the admissible arcs in order to optimise our coupling. For that, we will remove a value $\delta$ to all the valuations of the arcs between $X^*$ and $Y$-$Y^*$ and we will add this value $\delta$ to the valuations of the arcs between $X$-$X^*$ and $Y^*$. On the valuations between $X^*$ and $Y^*$, one adds and one subtracts $\delta$. All these operations are summarised in the figure 3.3.

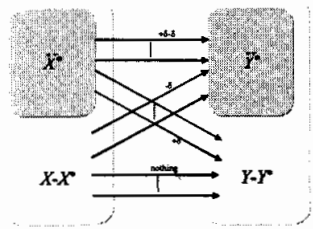

**figure 3.3** update of admissible arcs

This value $\delta$ is the minimum value (minimum belief) on the set of arcs which link $X^*$ and $Y$-$Y^*$:

$$\delta = \min(M_{i,j}(X^* \cap (Y - Y^*)))$$

We reiterate these operations of **coupling, improving chains search, marking** and **admissible arcs modification** until we obtain a maximal coupling.

### 3.2 Maximisation of the decision belief

With this assignment algorithm, we have an optimal decision in the sense of the maximisation of the sum of belief. But this algorithm is based on the processing of a square matrix, otherwise we add fictitious elements in order to have an exhaustive coupling, that is a coupling where each perceived object is affected to a known object. Each relation with a virtual object is valued with a belief equal to **0**.

For now, our decision is constituted of real or fictitious objects. In order to obtain our final decision, first we will take out these elements, then we will remove the assignments which belief is lower than the belief $m_{i,j}(*)$ associated with nothing. In fact, this second filtering enables us to use information on the unknown not used in the coupling algorithm. These two filterings are summarised by the following equation:

$$x_{i,j} = \begin{cases} 1 & if \quad m_{i,j}(Y_i) > \max_{i \leq |X|, j \leq |Y|}(m_{,j}(*), m_{i,\cdot}(*)) \\ 0 & else \end{cases}$$

$x_{i,j}$ represents the relation between $X_i$ and $Y_j$, this relation is validated if $x_{i,j}=1$ and is rejected if not.

### 3.3 Quantification of decision confidence

The cost that we will calculate from the sum of the beliefs will enable us to quantify the confidence we have in our decision. To say that a decision is " *the best* " is good but not sufficient. It is necessary to be able to quantify this concept of " *better* ". If in a case, the cost is .4 and that in a second case, we have a cost of .9, we will be certain in both cases that the two decisions are the best (the two cases are obviously independent). However we will tend to give more confidence to the second decision because this one reflects a greater reliability on association.

This confidence can be obtained by using the cardinality of the coupling which gives us maximum confidence and by using the beliefs. The coefficient thus obtained represents the percentage of confidence we have in our decision. Knowing the cardinal of our association, we know that, if we have a maximum confidence on all associations (we do not have any unknown factor then), the cost associated on our decision will be equal to this cardinal. This means that the belief on each association is equal to 1. Confidence we have on our decision is then:

$$\Psi = \frac{\sum_{i=1}^{X_n} \sum_{j=1}^{Y_m} C_{ij} \cdot x_{ij}}{\min(|X_n|, |Y_m|)}$$

$C_{ij}$ represents the belief that the object $X_i$ is in relation to the object $Y_j$.

$x_{i,j}$ represents the assignment or not of the object $X_i$ with the object $Y_j$.

## 4   Generation of the sets of masses

One of the difficulties for the implementation of this theory is the creation of the set of the initial masses. To generate them, we must firstly use a distance measurement that quantifies the similarity between our perceived objects and our known objects, and secondly we need an operator that generates our mass set from our similarity index. According to the model of representation for the used information, this index is computed with the distance of Mahalanobis in the statistical representation framework or, in the possibilist framework, with the possibilist index or the index of Jacquard [DUBO 88].

We studied an index of similarity from a representation of the objects by fuzzy quantities [GRUY 98b]. The support of a fuzzy quantity represents the inaccuracy around measurement and the height, its uncertainty.

The index of similarity (or agreement) quantifies, by a geometrical approach, the agreement between two fuzzy quantities that are either symmetrical, asymmetrical, normalised or sub-normalised. We give here two versions of this distance measurement.

**Similarity measurement for 1 dimension objects:**

$$X_i \cap Y_j = \frac{(a+b) \cdot h(X_i \cap Y_j)}{2}$$

$$X_i = \frac{(c+d) \cdot h(X_i)}{2}$$

$$\mathfrak{I}_s = \frac{X_i \cap Y_j}{X_i}\bigg|_{h(X_i)=1} = \frac{(a+b) \cdot h}{(c+d)}$$

**Similarity measurement for 2 dimensions objects:**

$$X_i \cap Y_j = \frac{h(X_i \cap Y_j)}{6} \cdot [c \cdot (2 \cdot b + b_2) + c_2 \cdot (2 \cdot b_2 + b)]$$

$$X_i = \frac{h(X_i)}{6} \cdot [i \cdot (2 \cdot a + a_2) + i_2 \cdot (2 \cdot a_2 + a)]$$

$$\mathfrak{I}_v = \frac{X_i \cap Y_j}{X_i}\bigg|_{h(X_i)=1} = \frac{h(X_i \cap Y) \cdot [c \cdot (2 \cdot b + b_2) + c_2 \cdot (2 \cdot b_2 + b)]}{[i \cdot (2 \cdot a + a_2) + i_2 \cdot (2 \cdot a_2 + a)]}$$

This index quantifies the intersection between the known object and the perceived object. It is normalised by the projection of the fuzzy measurement, which would certainty be equal to 1. This index indeed makes it possible to take into account the uncertainty and the inaccuracy of the objects (perceived or known).

The figures 4-1 a) and b) show the behaviour of this agreement operator (2D) when we apply a translation motion on the perceived object on his axis and when we increase the uncertainty on the known object. We see the influence of a strong inaccuracy of our fuzzy measurement on the computation of the measurement of similarity. Indeed, we observe that the value of the agreement index is never equal to 1 even if the certainty of the perceived object and the certainty of the known object are both maximum.

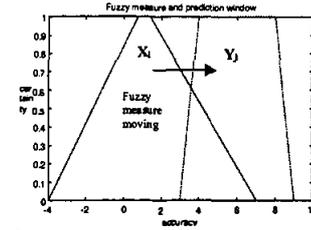

**Figure 4.1 a)** fuzzy quantities: perceived object and known object

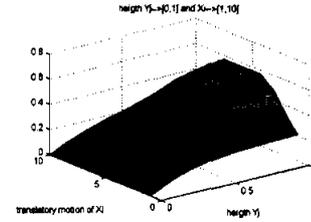

**Figure 4.1 b)** Concordance index behaviour

Using this index, we can generate our mass sets, but for that, it is necessary to find an adapted operator. Several works were already lead by [APPR 97], [DENO 95] or [ROMB 97]. Often these operators are well suited only to particular cases. The most traditional operators are based either on exponential (adapted to classification), or on probabilities which makes possible to handle the information modelled by the Gaussian. In our case, we want masses within [0,1] with the mass equal to 0 in the case of a total discordance and the mass equal to 1 for sources in total agreement. One wants also to be able to take into account the reliability of the information sources. This led us to develop this set of functions:

$$d_{i,j} = \pi \cdot (2 \cdot (1 - \mathfrak{I}_v) - 1)$$

$$m_{i,j}(Y_j) = \alpha_0 \cdot \left(1 - \left(\frac{\sin(\frac{d_{i,j}}{2}) + 1}{2}\right)\right)$$

$$m_{i,j}(\overline{Y}_j) = \alpha_0 \cdot \left(\frac{\sin(\frac{d_{i,j}}{2}) + 1}{2}\right)$$

$$m_{i,j}(\Theta_{i,j}) = 1 - \alpha_0$$

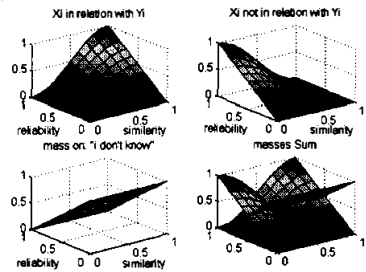

**Figure 4.2** Generation of the masses set

The coefficient $\alpha_0$ represents the reliability of information sources. The masses set thus generated has the properties of the belief theory and reflects the initial beliefs on the hypotheses of the frame of discernment.

## 5 Example

In this part, we will show a representative example of the operating mode of this multi-object association algorithm with ambiguity removal. The purposes of this example will be first to handle appearances and disappearances of objects or false alarms and secondly to maximise the sum of the beliefs on our association and thus to obtain "*the best*" decision.

This example simulates a scenario that includes three perceived objects and four known objects (fig 5.1). One object is a 1D information (distance). In the three perceived objects, we have a false alarm or the appearance of an object. In the four known objects, we have an object that probably has just disappeared. The sets of masses for each set of relations between a perceived object and the known objects are given below.

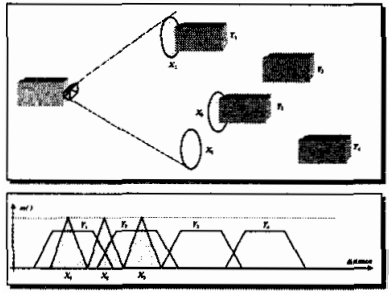

**figure 5.1** Our road situation corresponding to our scenario

Set of mass associated to $X_1$

$m_{1,1}(Y_1)=.8 \quad m_{1,2}(Y_2)=.5 \quad m_{1,3}(Y_3)=.1 \quad m_{1,4}(Y_4)=0$
$m_{1,1}(\overline{Y_1})=.1 \quad m_{1,2}(\overline{Y_2})=.4 \quad m_{1,3}(\overline{Y_3})=.8 \quad m_{1,4}(\overline{Y_4})=.9$
$m_{1,1}(\Theta_{1,1})=.1 \quad m_{1,2}(\Theta_{1,2})=.1 \quad m_{1,3}(\Theta_{1,3})=.1 \quad m_{1,4}(\Theta_{1,4})=.1$

Set of mass associated to $X_2$

$m_{2,1}(Y_1)=.5 \quad m_{2,2}(Y_2)=.5 \quad m_{2,3}(Y_3)=.1 \quad m_{2,4}(Y_4)=0$
$m_{2,1}(\overline{Y_1})=.1 \quad m_{2,2}(\overline{Y_2})=.1 \quad m_{2,3}(\overline{Y_3})=.7 \quad m_{2,4}(\overline{Y_4})=.9$
$m_{2,1}(\Theta_{2,1})=.4 \quad m_{2,2}(\Theta_{2,2})=.4 \quad m_{2,3}(\Theta_{2,3})=.2 \quad m_{2,4}(\Theta_{2,4})=.1$

Set of mass associated to $X_3$

$m_{3,1}(Y_1)=.4 \quad m_{3,2}(Y_2)=.8 \quad m_{3,3}(Y_3)=.1 \quad m_{3,4}(Y_4)=0$
$m_{3,1}(\overline{Y_1})=.1 \quad m_{3,2}(\overline{Y_2})=.1 \quad m_{3,3}(\overline{Y_3})=.6 \quad m_{3,4}(\overline{Y_4})=.9$
$m_{3,1}(\Theta_{3,1})=.5 \quad m_{3,2}(\Theta_{3,2})=.1 \quad m_{3,3}(\Theta_{3,3})=.3 \quad m_{3,4}(\Theta_{3,4})=.1$

From these sets of masses and by using the equations given in part 2.3 we obtain a new masses set represented in two matrices of beliefs $M^{\sigma}_{i,\cdot}$ and $M^{\sigma}_{\cdot,j}$. The first gives the belief of the relations between the perceived objects and the known objects and the second the belief between the known objects and the perceived objects.

| $M^{\sigma}_{i,\cdot}$ | $Y_1$ | $Y_2$ | $Y_3$ | $Y_4$ | * | $\Theta$ |
|---|---|---|---|---|---|---|
| $X_1$ | 0.6545 | 0.1636 | 0.0182 | 0 | 0.0524 | 0.1113 |
| $X_2$ | 0.3214 | 0.3214 | 0.0357 | 0 | 0.0090 | 0.3124 |
| $X_3$ | 0.1154 | 0.6923 | 0.0192 | 0 | 0.0087 | 0.1644 |

| $M^{\sigma}_{\cdot,j}$ | $X_1$ | $X_2$ | $X_3$ | * | $\Theta$ |
|---|---|---|---|---|---|
| $Y_1$ | 0.6000 | 0.1500 | 0.1000 | 0.0025 | 0.1475 |
| $Y_2$ | 0.1429 | 0.1429 | 0.5714 | 0.0114 | 0.1314 |
| $Y_3$ | 0.0833 | 0.0833 | 0.0833 | 0.3457 | 0.4043 |
| $Y_4$ | 0 | 0 | 0 | 0.7290 | 0.2710 |

For each one of these matrices, we obtained a decision by using the maximum of belief. The first decision obtained on the matrix $M^{\sigma}_{i,\cdot}$ is:

$X_1$ is in relation to $Y_1$     $X_2$ is in relation to $Y_1$
$X_2$ is in relation to $Y_2$     $X_3$ is in relation to $Y_2$

And the second decision with the matrix gives us:

$Y_1$ is in relation to $X_1$     $Y_2$ is in relation to $X_3$
$Y_3$ is in relation to $\Theta$     $Y_4$ is in relation to *

We can deduce from the first decision we have a conflict on the object to associate with $X_2$. The second decision shows firstly a confirmation of the association of $X_1$ with $Y_1$ and $X_3$ with $Y_2$, secondly a lake of knowledge on the association of $Y_3$, and thirdly an association of the fourth object $Y_4$ with nothing. In order to solve this conflict, we will use the algorithm of ambiguity removal on the new matrix resulting from the combination of our two belief matrices.

| $M^{\sigma}_{i,j}$ | $Y_1$ | $Y_2$ | $Y_3$ | $Y_4$ |
|---|---|---|---|---|
| $X_1$ | 0.3927 | 0.0234 | 0.0015 | 0 |
| $X_2$ | 0.0482 | 0.0459 | 0.0030 | 0 |
| $X_3$ | 0.0115 | 0.3956 | 0.0016 | 0 |

In order to get a square matrix $M^{\sigma}_{i,j}$, we add a virtual perceived object $X_4$ with for each one of its relations with the known objects a belief of 0. To reveal the admissible arcs in our matrix (given by $\overline{C}_{ij}=1-C_{ij}$), we use the cost matrix $\overline{M^{\sigma}_{i,j}}$ with a complement to 1. The result of this assignment algorithm gives the following association matrix:

| $X_{i,j}$ | $Y_1$ | $Y_2$ | $Y_3$ | $Y_4$ |
|---|---|---|---|---|
| $X_1$ | 1 | 0 | 0 | 0 |
| $X_2$ | 0 | 0 | 1 | 0 |
| $X_3$ | 0 | 1 | 0 | 0 |
| $X_4$ | 0 | 0 | 0 | 1 |

By applying our filtering, we can immediately eliminate the association with the virtual objects (criterion on the cardinality: $i \leq |X|$ and $j \leq |Y|$). In our case, $X_4$ is our virtual object.

Then we will use information on the belief $m_{i,j}(*)$, that is the information on the fact that an object is affected with nothing, to filter the remainder of the objects.

We have $m_{2,3}(Y_3) < \max(m_{\cdot,3}(*), m_{2,\cdot}(*))$ thus $X_2$ is associated with "*nothing*". We obtain then as a final decision $X_1$ associated with $Y_1$, $X_2$ associated with nothing and $X_3$ associated with $Y_2$. This enables us to build the following assignment matrix:

| $X_{i,j}$ | $Y_1$ | $Y_2$ | $Y_3$ | $Y_4$ | * |
|---|---|---|---|---|---|
| $X_1$ | 1 | 0 | 0 | 0 | 0 |
| $X_2$ | 0 | 0 | 0 | 0 | 1 |
| $X_3$ | 0 | 1 | 0 | 0 | 0 |
| $X_4$ | 0 | 0 | 0 | 0 | 1 |

## 6. Temporal propagation of the objects

This association is part of the design of a wider algorithm enabling us to do multi-target tracking and dynamic environment cartography. Moreover, the management of these appearances and disappearances make it possible to propagate virtual objects through their prediction. These virtual objects then have an uncertainty in time. When this uncertainty becomes too great, the object disappears.
This object propagation reduces the effect of awkward events such as objects crossing, measure deterioration due to weather conditions, or temporary sensors degradation (information missing or false alarms).
This multi-object tracking algorithm avoids some problems encountered by other algorithms of the same kind like the PDAF, which is not adjusted to targets crossing, the JPDAF, that takes into account a fixed number of targets and doesn't initialise new tracks, or the MHT that has combinatorial problems [SHAL 92].

## 7 Conclusion and future works

This algorithm enables us to combine the opinions we have on relations between objects, we can take into account the inaccuracy and uncertainty on all measurements and predictions. We are also able, by using these fuzzy models of data, to generate sets of masses representative of the current situation. Moreover, this combination has the advantage to be associative and commutative, which is difficult to obtain with the majority of data fusion operators. By generalising Dempster combination rule, we also showed that it is possible to reduce the complexity of this combination and to make it independent of the recurrence.
With the assignment algorithm, we showed we give a decision that we can affirm to be optimal, that is the "*best*". We are also able to quantify the confidence we have in this decision. This algorithm must be integrated in a vehicle perception system in order to carry out the cartography of the dynamic environment surrounding the vehicle to characterise the current road situation.
We then focus on the initialisation stage of this algorithm we wanted as simple and as automatic as possible. It will not depend on constraining heuristic parameters complicated to implement. It will not suffer from constraints due to human interventions.
This multi-object tracking algorithm must be implemented soon on the laboratory vehicle (**STRADA**). This vehicle will be equipped with many sensors: cameras, laser telemeter and proprioceptive sensors.